\pdfoutput=1
\documentclass[11pt]{article}

\usepackage{ACL2023}

\usepackage{times}
\usepackage{latexsym}

\usepackage[T1]{fontenc}
\usepackage[utf8]{inputenc}

\usepackage{microtype}

\usepackage{inconsolata}

\usepackage{mathrsfs}
\usepackage{amsmath}
\usepackage{amssymb}
\usepackage{bbm}
\usepackage{graphicx}
\usepackage{array,multirow}
\usepackage{float}
\usepackage{threeparttable}
\usepackage{enumerate}
\usepackage{comment}
\usepackage{subcaption}
\usepackage{tabularx}
\usepackage{enumitem}
\usepackage{algorithmic}
\usepackage[linesnumbered,ruled,vlined]{algorithm2e}
\SetKwInput{KwInput}{Input}                % Set the Input
\SetKwInput{KwOutput}{Output}              % set the Output
\SetKwInput{KwParameter}{Parameter}        % set the Output

\usepackage{tikz}
\newcommand*\circled[1]{\tikz[baseline=(char.base)]{
  \node[shape=circle,draw,inner sep=1pt] (char) {#1};}}

\title{A Scalable and Adaptive System to Infer the Industry Sectors of Companies: Prompt + Model Tuning of Generative Language Models}

\author{
  Lele Cao\thanks{~~Corresponding author.}~{\normalfont ,}\
  Vilhelm von Ehrenheim{\normalfont ,}\
  Astrid Berghult{\normalfont ,}\
  Cecilia Henje{\normalfont ,}\
  Richard Anselmo Stahl{\normalfont ,} \\
  {\bf Joar Wandborg},\
  {\bf Sebastian Stan},\
  {\bf Armin Catovic},\
  {\bf Erik Ferm} \and
  {\bf Hannes Ingelhag} \\
 Motherbrain, EQT Group, Stockholm, Sweden \\
\texttt{\small\{lele.cao, vilhelm.vonehrenheim, astrid.berghult, cecilia.henje, richard.stahl\}@eqtpartners.com} \\
\texttt{\small\{joar.wandborg, sebastian.stan, armin.catovic, erik.ferm, hannes.ingelhag\}@eqtpartners.com}
}

\begin{document}
\maketitle
\begin{abstract}
%\small
The Private Equity (PE) firms operate investment funds by acquiring and managing companies to achieve a high return upon selling. Many PE funds are thematic, meaning investment professionals aim to identify trends by covering as many industry sectors as possible, and picking promising companies within these sectors. So, inferring sectors for companies is critical to the success of thematic PE funds. In this work, we standardize the sector framework and discuss the typical challenges; we then introduce our sector inference system addressing these challenges. Specifically, our system is built on a medium-sized generative language model, finetuned with a prompt + model tuning procedure. The deployed model demonstrates a superior performance than the common baselines. The system has been serving many PE professionals for over a year, showing great scalability to data volume and adaptability to any change in sector framework and/or annotation.

\end{abstract}

\section{Introduction}

Private Equity (PE), as a fast-growing branch of the investment industry, operates investment funds on behalf of institutional and accredited investors by acquiring and managing companies before selling them to achieve high, risk adjusted returns.
% PE funds may acquire majority shares of private or public companies, or invest in buyouts as part of a consortium.
% The PE industry has grown rapidly; it tends to become especially popular when stock prices are high and interest rates low.
The common PE investment strategies, according to \cite{block2019private}, 
% include Venture Capital (VC), Growth Capital (GC), and Leveraged Buyouts (LBO).
include venture capital, growth capital, and leveraged buyouts.
The majority of PE funds strive to be ``thematic'' \cite{berube2014indexes}, aiming to identify macro-level trends by covering a variety of relevant sectors and picking promising companies within these sectors.
In order to do that, any company should be put into a sector that best describes its main business activity.
The sectors are often defined hierarchically (cf.~the sector framework in Section~\ref{sec:sf}), where the sectors higher up in the hierarchy tend to have a broader scope (hence usually fewer in number) and be more stable, while the ones lower down (a.k.a. ``industries'') are more fine-grained and prone to change.
A well-defined sector framework enables investment professionals to conduct a deeper analysis of the economy within each individual sector.

\begin{figure}
    \centering
    \includegraphics[width=\columnwidth]{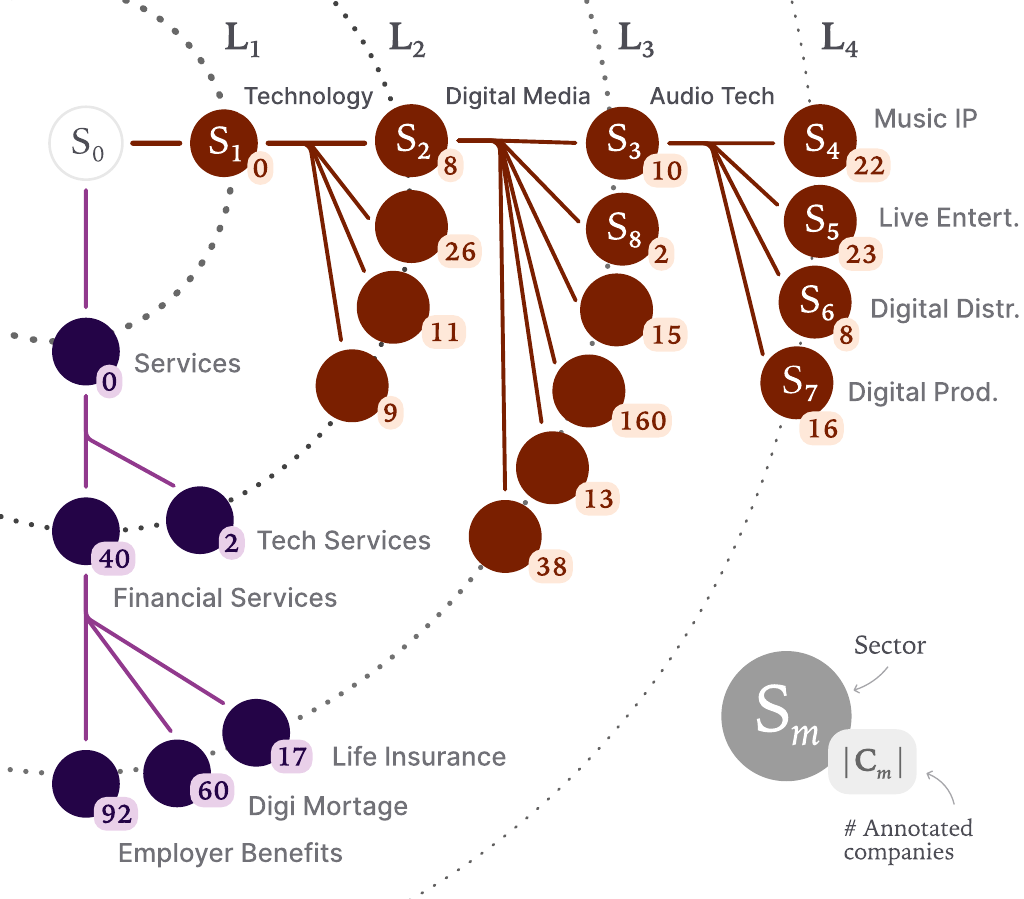}
    \caption{A PE sector framework defined as a tree with a depth $L$=4. Each non-root node represents a sector (i.e., $\mathsf{s}_1 \sim \mathsf{s}_M$) that is numbered in a depth-first order. The integer attached to the $m$-th sector/node indicates the number of companies $|\mathsf{C}_m|$ annotated for $\mathsf{s}_m$.}
    \label{fig:sf}
\end{figure}

There are currently hundreds of millions of companies worldwide, and thousands of new companies are founded daily.
Realistically, human professionals can only evaluate a limited number of companies to determine their belonging sectors.
In order to significantly increase the coverage of sector mapping, practitioners have begun resorting to predictive systems to infer the belonging sectors of companies.
Due to the reasons discussed in Section~\ref{sec:sf}, there has not been any effective system that is generic enough to drive the wide adoption in PE operations.
In this paper, we standardize the sector framework and discuss the typical challenges; we then introduce our sector inference system addressing these challenges.
Our system has been successfully serving hundreds of PE professionals for over a year.
The highlight is three fold:
\begin{itemize}
\item We propose to co-tune the PLM (pretrained language model) starting from a later stage of prompt tuning, attempting to leverage the capability of medium-sized PLMs to an extreme using scarce annotation.
\item We implement an autonomous system, which effectively handles the dynamic sector framework, evolving annotation, data imbalance, noisy features, and high inference volume.

\item We experimentally show the superior performance of our approach in comparison to the common baselines, and justify many design choices such as model paradigm and size.
\end{itemize}

\section{The Problem and Challenges}
\label{sec:sf}

Let $\mathsf{c}_n$ denote the $n$-th company ($n\!=\!1,2,\ldots,N$) in the scope of a PE firm; the total number of companies $N$ usually reaches the order of millions.
Most of the time, PE professionals maintain a hierarchical sector framework containing $M$ different sectors represented as nodes ($\mathsf{s}_1\!\sim\!\mathsf{s}_M$) in a tree with $L$ layers, as illustrated in Figure~\ref{fig:sf}.
In practice, the value of $L$ is mostly less than 4, and the total number of sectors (i.e.,~$M$) tracked by a large PE firm may reach up to a few hundred.
The problem is {\bf how to assign each company $\mathsf{c}_n$ to the most relevant sector $\mathsf{s}_m$}.
Solving such a problem requires addressing several {\bf challenges} (abbreviated as {\bf\small Chall.}) that will be discussed below.

{\bf Chall.1}: scarce, imbalanced and evolving annotation.
One might notice there are some public datasets such as G2 and Pitchbook\footnote{\url{https://www.g2.com} and \url{https://pitchbook.com}} that contain sector annotations, i.e., $\mathsf{c}_n\!\to\!\mathsf{s}_m$.
In reality, they can not be directly used to train the sector inferring model, which is the consequence of two main facts:
(1) PE firms almost always maintain their own version of sector framework that are drastically different from the ones from public datasets.
(2) PE funds may annotate companies differently; for example, Klarna\footnote{\url{https://www.klarna.com}} might fall into any sector of {\it payment method}, {\it digital bank} and {\it financial service} depending on the preference of investment professionals or the fund specifications.
%\verb|payment_method|, \verb|digital_bank| and \verb|financial_service|.
To that end, we allow professionals to select a sector for any company via the investment platform developed in-house.
Formally, we use $\mathsf{C}_m$ to denote the set of companies annotated for sector $\mathsf{s}_m$,
and the total number of companies in $\mathsf{C}_m$ is $|\mathsf{C}_m|$;
taking node \circled{$\mathsf{s}_3$} in Figure~\ref{fig:sf} for example, its subscript {\tiny\boxed{10}} contains the value of $|\mathsf{C}_3|$, i.e.,~$|\mathsf{C}_3|\!=\!10$.
In reality, the sector annotation is scarce (an intrinsic limitation of manual annotation), imbalanced ($|\mathsf{C}_m|$ can vary greatly among sectors) and ever-evolving (the mapping $\mathsf{c}_n\!\to\!\mathsf{s}_m$ may change frequently).

{\bf Chall.2}: dynamic sector framework with varying granularity.
Due to shifting market trends, the sector framework is rarely fixed for extended periods of time.
Instead, the sector framework is really a dynamic one, where one of the three changes\footnote{Note that changing the definition of an existing sector is achieved by altering the associated company annotations; and merging/splitting existing sector(s) can be done via combining operations of adding and removing sector(s). Currently, only a system superuser can modify the sector framework through backend configuration files. However, our future plans involve facilitating this process via a web-based graphical user interface (GUI) integrated to EQT's Motherbrain platform -- \url{https://eqtgroup.com/motherbrain}.} can occur: adding new layers, adding new nodes, and removing nodes.
Another observation is that PE professionals will pick concepts they think are important and define them as sectors, leading to sectors with varying granularity even on the same layer. 
For instance, a sector could be anything from a new technology (e.g.,~block chain), an environmental concern (e.g.,~water shortage), to an emerging market demand (e.g.,~Coronavirus test).

{\bf Chall.3}: availability and quality of features.
Intuitively, the most informative feature is probably the textual description about a company, which can be gathered from various data sources such as Pitchbook and Crunchbase\footnote{\url{https://www.crunchbase.com}}.
Given an example description ``{\it \small We develop security analytical tools to identify web-app vulnerabilities. Contact us for a demo of our award-winning product}'', one could guess a ``{\it\small cyber security}'' sector just by reading the first sentence, yet many texts look more like the second sentence, which severely lacks context.
Moreover, a significant number of companies simply do not have textual descriptions available from popular data sources\footnote{\citet{cao2022using} present a summary of data sources.}.

{\bf Chall.4}: high inference frequency and volume.
As soon as the textual feature of a company is changed, we need to re-infer its sector.
Besides, any change around the sector framework or company annotation may trigger model update, which requires a re-inference for all $N$ companies. 
With the current data volume in our data warehouse, the daily amount of re-inference can easily exceed 100 million, which may grow into a bottleneck.
%For conciseness, we abbreviate {\bf Challenge} as {\bf Chall.}

\section{The Core Model}
\label{sec:model}
Inferring the industry sector of companies can be naturally addressed by a supervised NLP approach, where we {\bf input} the textual description of a company (denoted as $\mathbf{c}$), and {\bf output} a sector $\mathsf{s}$ based on a $\boldsymbol{\theta}$-parameterized model $P(\mathsf{s}|\mathbf{c};\boldsymbol{\theta})$;
note that we omit the subscripts $n$ and $m$ hereafter for the sake of simplicity. 
To find the optimal $\boldsymbol{\theta}$, we use the annotated mappings $\mathbf{c}\!\to\!\mathsf{s}$ to fit this conditional probability.
The prediction target $\mathsf{s}$ can be either raw text (e.g.,~``{\it \small cyber security}'') or the encoded $M$-dim one hot vector, where the former is a {\bf generative} approach and the latter is {\bf discriminative}.
It is crucial to highlight that generative methods offer two primary advantages over discriminative techniques (such as supervised classification): (1) generative models are capable of predicting sectors beyond those predefined, and (2) since these models output natural words, they can more effectively harness pre-learned knowledge in LM, thus avoiding overfitting on smaller training datasets.

\begin{figure}
    \centering
    \includegraphics[width=.46\textwidth]{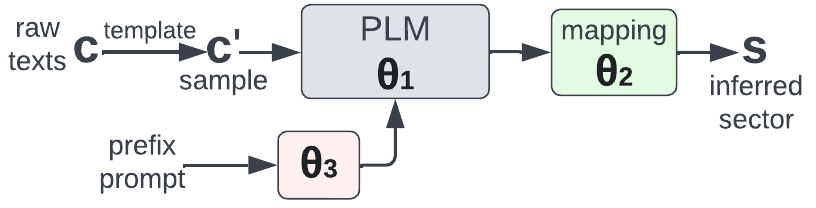}
    \caption{Three paradigms of generative NLP models: prompting $P(\mathsf{s}|\mathbf{c};\boldsymbol{\theta}_2)$, model tuning $P(\mathsf{s}|\mathbf{c};\boldsymbol{\theta}_1,\boldsymbol{\theta}_2)$, and prompt tuning $P(\mathsf{s}|\mathbf{c};\boldsymbol{\theta}_2,\boldsymbol{\theta}_3)$.}
    \label{fig:3m}
\end{figure}

Nowadays, generative approaches dominate the domains of computer vision (e.g.,~Stable Diffusion by \citealp{rombach2022high}) and NLP (e.g.,~GPT-3 \cite{brown2020language} and GPT-4 \cite{openai2023gpt4}).
Particularly, the language model (LM) is often pretrained following a generative approach, such as predicting the masked words.
To address {\bf\small Chall.1}{\small\&\bf 2}, we need to exploit the capability of a pretrained LM (PLM). 
We start with designing a template for samples: 
%``{\it \small company: \verb|[NAME]| description: \verb|[SHORT_TEXT]| tags: \verb|[TAGS]| details: \verb|[LONG_TEXT]| sector: \verb|[MASK]|}''
\begin{quote}
\small \verb|[NAME]|, concerns \verb|[TAGS]|, is \verb|[|$\mathbf{c}$\verb|]|. Sector: \verb|[|$\mathbf{s}$\verb|]|.
\end{quote}
For a certain company, {\small \verb|[NAME]|} is its legal name, 
{\small \verb|[TAGS]|} is the concatenated tags/keywords\footnote{Many data sources, such as Pitbook and Crunchbase, have some keywords tagged for each company.} that are added to address {\bf\small Chall.3}.
For example, the filled input for company ``Klarna'' may look like
\begin{quote}
\it \small Klarna Bank AB, concerns buy-now-pay-later and shopping, is an online payment platform designed to facilitate cashless payments. Sector: \verb|[|$\mathbf{s}$\verb|]|.
\end{quote}
As the prediction target, {\small \verb|[|$\mathbf{s}$\verb|]|} remains unreplaced, thus it is an unanswered sample. 
Model optimization essentially attempts to make the predicted {\small \verb|[|$\mathbf{s}$\verb|]|} closer to the annotated sector text, and in this example $\mathbf{s}$ = ``{\it financial service}''.
We use $\mathbf{c}'$ to denote the filled sample for company $\mathsf{c}$.

\subsection{Prompt and Model Tuning}
\label{sec:prompt_model_tuning}
Despite minor differences, the generative NLP models largely adhere to one of three paradigms: prompting, model tuning, or prompt tuning. 
Seen from Figure~\ref{fig:3m}, {\bf prompting} \cite{liu2021pre} freezes the PLM weights $\boldsymbol{\theta}_1$ while learning a mapping function (parameterized with $\boldsymbol{\theta}_2$) to transform the raw PLM output into the sector space.
{\bf Model tuning} allows finetuning $\boldsymbol{\theta}_1$, which is the de facto way of leveraging
large PLM for downstream tasks \cite{li-liang-2021-prefix}.
{\bf Prompt tuning} prepends some soft prompts, which are essentially learnable virtual tokens, into the input sequence $\mathbf{c}'$ and only trains them (corresponding to parameter $\boldsymbol{\theta}_3$) while keeping $\boldsymbol{\theta}_1$ fixed \cite{su2022transferability}.
PLM can have billions of parameters making model tuning paradigm expensive, while the prompt-based approach \cite{liu2021pre} has only thousands of tunable parameters \cite{lester-etal-2021-power}.

\begin{figure}[t!]
\centering
\subcaptionbox{\footnotesize Average precision.\label{fig:paradigm_size_precision}}
{\includegraphics[height=5.6cm]{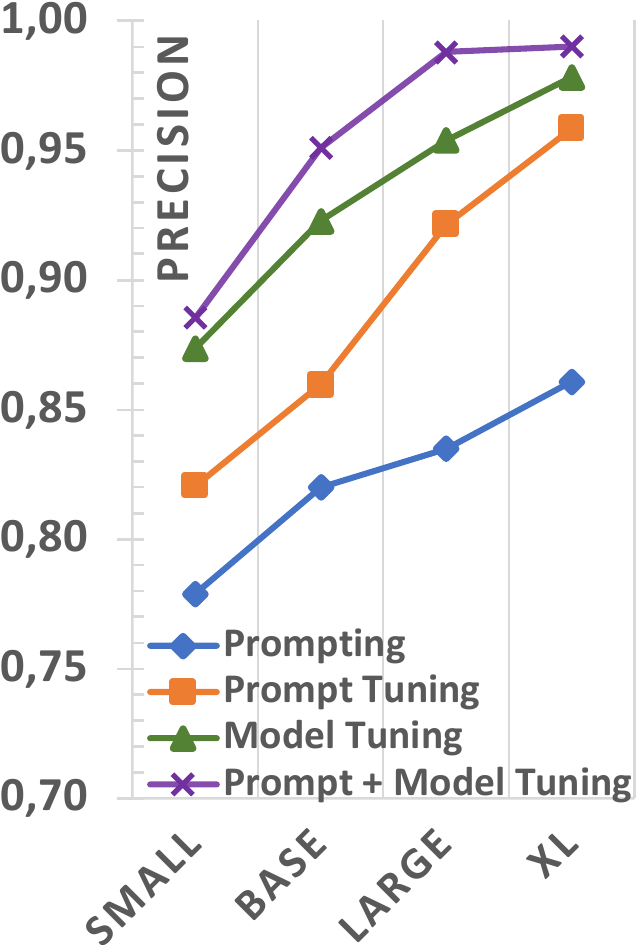}}
\hspace{1pt}
\subcaptionbox{\footnotesize Average recall.\label{fig:paradigm_size_recall}}
{\includegraphics[height=5.6cm]{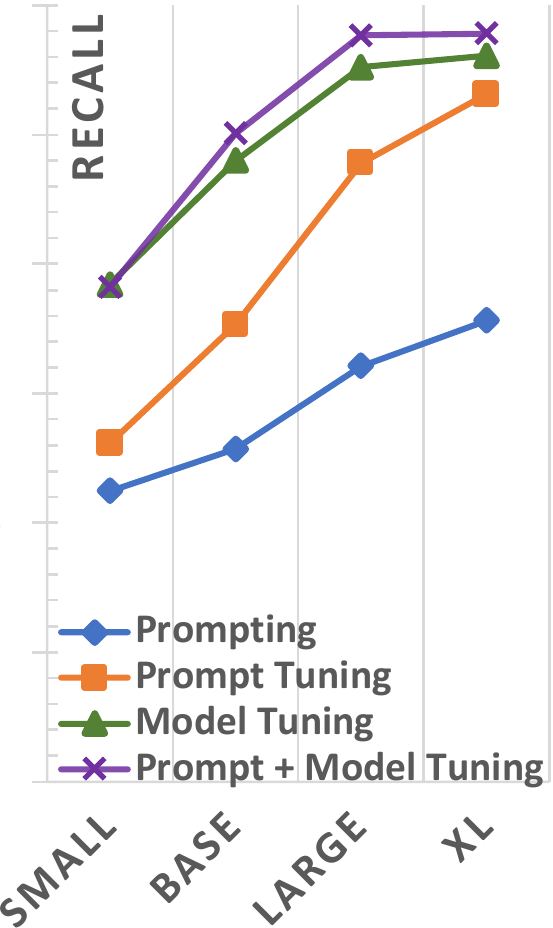}}
\caption{\label{fig:paradigm_size}Performance comparison over different model paradigms (legend) and sizes (x-axis): the validation (a) precision and (b) recall averaged over 84 sectors.}
\end{figure}

Following the generative ``text-to-text'' T5 PLM \cite{JMLR:v21:20-074}, we compared the performance of these paradigms towards the PE sector inferring task.
Figure~\ref{fig:paradigm_size} shows the average precision and recall in relation to different paradigms and model sizes (\verb|Small|, \verb|Base|, \verb|Large| and \verb|XL|)\footnote{We did not manage to experiment the \texttt{XXL} T5 model due to our restriction of computing and human resources. The T5 PLMs can be found in \url{https://huggingface.co/google}.}.
We observe that model tuning of T5 achieves stronger performance than prompting and prompt tuning.
Prompt tuning catches up with model tuning as model size increases, which coincide the conclusion drawn by 
\citet{lester-etal-2021-power}.
Intuitively, the label scarcity ({\bf\small Chall.1}) and varying granularity of sector framework ({\bf\small Chall.2}) could be better addressed by prompt tuning, since it is supposed to keep the learned knowledge in PLM untouched;
meanwhile, we also want to replicate the superior performance of model tuning when using a smaller model.
To that end, we propose a model $P(\mathsf{s}|\mathbf{c};\boldsymbol{\theta}_1,\boldsymbol{\theta}_2,\boldsymbol{\theta}_3)$ that carries out $t'$ steps of prompt tuning (only optimize $\boldsymbol{\theta}_2$ and $\boldsymbol{\theta}_3$) before jointly tune the PLM weights $\boldsymbol{\theta}_1$, as presented in Algorithm~\ref{algo:pmt}.
Seen from Figure~\ref{fig:paradigm_size}, this ``{\small Prompt + Model Tuning}'' approach outperforms all compared methods by a large margin, which is the case even when the PLM size is relatively small.

\begin{algorithm}[t]
\footnotesize
\DontPrintSemicolon
\KwInput{Sector annotations in the form of $\mathsf{c}\!\to\!\mathsf{s}$, a generative NLP model 
$P(\mathsf{s}|\mathbf{c};\boldsymbol{\theta}_1,\boldsymbol{\theta}_2,\boldsymbol{\theta}_3)$, PLM freezing steps $t'$, learning rates $\epsilon_1$ and $\epsilon_2$}
\KwOutput{The optimal parameters $\boldsymbol{\theta}^*_1$, $\boldsymbol{\theta}^*_2$ and $\boldsymbol{\theta}^*_3$}
    
Initialize $\boldsymbol{\theta}_1$ by loading the pretrained T5 model;

Initialize $\boldsymbol{\theta}_2$ and $\boldsymbol{\theta}_3$ randomly;

\For{$(t = 1; t \leq T; t++)$}{
    
    Sample a mini-batch from the annotations;

    Transform each $\mathsf{c}$ into a filled template $\mathbf{c}'$;

    Forward propagate $\mathbf{c}'$ to obtain the prediction $\hat{\mathsf{s}}$;

    Calculate the T5 cross entropy loss $\mathcal{L}(\hat{\mathsf{s}}, \mathsf{s})$;

    \If {$t \leq t'$}{

        $\epsilon=\epsilon_1$;
    }
    
    \Else{

        $\epsilon=\epsilon_2\;$ and 
        $\;\boldsymbol{\theta}_1\!:=\!\boldsymbol{\theta}_1\!-\!\epsilon\frac{\partial\mathcal{L}(\hat{\mathsf{s}}, \mathsf{s})}{\partial\boldsymbol{\theta}_1}$;

    }

    $\boldsymbol{\theta}_2\!:=\!\boldsymbol{\theta}_2\!-\!\epsilon\frac{\partial\mathcal{L}(\hat{\mathsf{s}}, \mathsf{s})}{\partial\boldsymbol{\theta}_2}\;$ and
    $\;\boldsymbol{\theta}_3\!:=\!\boldsymbol{\theta}_3\!-\!\epsilon\frac{\partial\mathcal{L}(\hat{\mathsf{s}}, \mathsf{s})}{\partial\boldsymbol{\theta}_3}$;
    
}

$\boldsymbol{\theta}^*_1=\boldsymbol{\theta}_1$, $\boldsymbol{\theta}^*_2=\boldsymbol{\theta}_2$ and $\boldsymbol{\theta}^*_3=\boldsymbol{\theta}_3$;

\textbf{return} $\boldsymbol{\theta}^*_1$, $\boldsymbol{\theta}^*_2$ and $\boldsymbol{\theta}^*_3$;

\caption{Prompt + model tuning}
\label{algo:pmt}
\end{algorithm}

According to Figure~\ref{fig:paradigm_size}, the performance of our approach increase with the size of PLM and plateau (>98\%) when reaching a ``\verb|Large|'' size.
Hence, we initialize our model with the T5-\verb|Large| PLM and train for $T\!=\!1\!\times\!10^6$ steps with a mini-batch size of 50.
The prompt tuning phase is trained for $t'\!=\!3\!\times\!10^3$ steps with a learning rate of $\epsilon_1\!=\!0.1$, where the first $1\!\times\!10^3$ steps utilize a linear learning rate warm-up \cite{goyal2017accurate}.
Afterwards, the joint prompt and model tuning begins with a warm-up of $1.5\!\times\!10^{3}$ steps until reaching a learning rate of $\epsilon_2=5\!\times\!10^{-3}$.
Checkpoints are selected via early stopping with respect to the validation accuracy.
All these hyper-parameters are determined by an empirical grid search, and the implementation is built upon OpenPrompt \cite{ding2021openprompt}.

\subsection{Annotation attribution}
\label{sec:aa}
\begin{figure}
    \centering
    \includegraphics[width=\columnwidth]{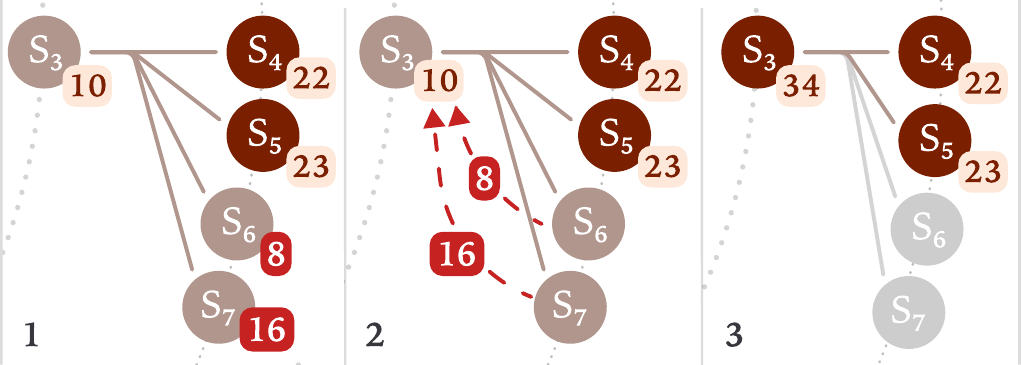}
    %\includesvg[inkscapelatex=false,width=\columnwidth]{figures/label_attribution.svg}
    \caption{Demonstration of annotation (label) attribution process using $\mathsf{s}_3$ sub-tree from Figure~\ref{fig:sf} as an example. Darker colored nodes are eligible for modeling.}
    \label{fig:la}
\end{figure}

We empirically regulate that only the sectors with at least 20 annotated companies can be included in the modeling, which implies that some sectors, such as $\mathsf{s}_3$, $\mathsf{s}_6$ and $\mathsf{s}_7$ in Figure~\ref{fig:sf}, are not eligible directly.
Since sector annotation is scarce ({\bf\small Chall.1}), we try to utilize every annotation to predict as much sectors as possible.
Subsequently, we run a depth-first (bottom-up) annotation attribution algorithm to collect the eligible sectors $\mathsf{s}_m$ and their annotated set of companies $\mathsf{C}_m$.
Figure~\ref{fig:la} demonstrates this procedure in three steps assuming the annotation attribution algorithm is currently processing the $\mathsf{s}_3$ sub-tree in Figure~\ref{fig:sf}.
%As exemplified in Figure~\ref{fig:la}, we use notation $\mathsf{s}_m\!\!:\!|\mathsf{C}_m|$ to represent an eligible sector $\mathsf{s}_m$ with $|\mathsf{C}_m|$ annotated companies.
%Suppose the algorithm is currently processing the $\mathsf{s}_3$ sub-tree in Figure~\ref{fig:sf}, 
Initially, only the child sectors $\mathsf{s}_4$ and $\mathsf{s}_5$ are eligible (cf.~\circled{$\mathsf{s}_4$}{\tiny\boxed{22}} and \circled{$\mathsf{s}_5$}{\tiny\boxed{23}} in Figure~\ref{fig:la}) because they have more than 20 annotated companies.
When it comes to $\mathsf{s}_6$ and $\mathsf{s}_7$, they have insufficient annotations, thus are not eligible.
However, their annotations will move up and contribute to the parent sector $\mathsf{s}_3$, enabling $\mathsf{s}_3$ to be included in the training dataset due to $|\mathsf{C}_3|$=10+8+16=34>20.

% Need to add $|\mathsf{C}_m|$ to the left of the Figure

Despite our best-effort annotation attribution procedure, it is possible that some sectors may still be excluded from training. However, in practice, the trained generative model is capable of producing sector names that are not within the eligible sector set. We believe this occurrence represents scenarios where the sectors are not covered by the labels, yet they are still significant in terms of their inherent business implications. This feature is particularly desirable as it facilitates better understanding and refinement of the sector framework.

\begin{algorithm}[t]
\footnotesize
\DontPrintSemicolon
\KwInput{The eligible sectors $\mathsf{s}_1,\ldots,\mathsf{s}_M$ and their corresponding company sets $\mathsf{C}_1,\ldots,\mathsf{C}_M$}
\KwOutput{The balanced company sets $\mathsf{C}'_1,\ldots,\mathsf{C}'_M$}

Initialize: $\mathsf{C}'_1=\mathsf{C}_1,\ldots,\mathsf{C}'_M=\mathsf{C}_M$; 

Calculate $\zeta=2\cdot\max\{|\mathsf{C}_1|,\ldots,|\mathsf{C}_M|\}$;

\For{$(m = 1; m \leq M; m++)$}{
    
    \For{each $\mathsf{c}$ in $\mathsf{C}_m$}{

        Augment $\mathsf{c}$ for $\lfloor\zeta/\mathsf{C}_M\rfloor\!-\!1$ times with EDA \cite{wei-zou-2019-eda}, producing set $\mathsf{c}'$;

        $\mathsf{C}'_m=\mathsf{C}'_m\cup\mathsf{c}'$;
    
    }

}

\textbf{return} $\mathsf{C}'_1,\ldots,\mathsf{C}'_M$;

\caption{Sample balancing via EDA}
\label{algo:sba}
\end{algorithm}

\subsection{Sample balancing via augmentation}
\label{sec:aug}
As a part of {\bf\small Chall.1} discussed in Section~\ref{sec:sf}, the value of $|\mathsf{C}_m|$ can vary from merely 20 all the way to a few hundred.
Thus, the aforementioned annotation attribution will produce a heavily imbalanced training dataset.
The overall idea is augmenting the samples for minority sectors to achieve inter-sector balance.
There is a whole spectrum of text augmentation methods: from rule-based to model-based techniques \cite{feng-etal-2021-survey}, from which we adopt the EDA (easy data augmentation) approach \cite{wei-zou-2019-eda} because of its simplicity and universality.
For individual {\small \verb|[NAME]|}, {\small \verb|[TAGS]|} and {\small \verb|[|$\mathbf{c}$\verb|]|} from our sample template, we perform synonym replacement, insertion, swapping and deletion at random choice with random intensity.
Algorithm~\ref{algo:sba} has the details of the entire balancing procedure.

\subsection{Performance analysis}
As of December 2022, there are 84 eligible sectors after the annotation attribution procedure as introduced in Section~\ref{sec:aa}.
We collect all samples manually annotated under one of these 84 sectors, thereby creating a dataset that exhibits imbalance in terms of the number of samples annotated for each sector.
The dataset is then balanced via the augmentation procedure introduced in Section~\ref{sec:aug}. 
This results in a final dataset containing 7,260 samples, where each sector has $\sim$86 annotated samples in average.
We reserve 15\% of the dataset for validation and report the accuracy of different baselines in Table~\ref{tab:perform-comp}.
Our approach (i.e.,~``Prompt + Model Tuning'') manages to achieve an accuracy of over 80\% on the validation set. 
In contrast, its discriminative counterpart (cf.~Section~\ref{sec:rw}), which employs an $M$-way classification output head, achieves only 70\% accuracy (largely on par with prompt tuning), likely due to the scarcity of labels.

\begin{table}[t]
\addtolength{\tabcolsep}{-2pt}
\centering
\begin{tabular}{l|r}
\hline
\textbf{Model} & Accuracy (\%) \\
\hline
$M$-Way Classification & $70.02$ \\ 
Prompting & $64.63$ \\ 
Prompt Tuning & $70.91$ \\ 
Model Tuning & $76.44$ \\ 
Prompt + Model Tuning (Ours) & $\text{\bf 80.25}$\\
\hline
\end{tabular}
\caption{Performance comparison of various baselines, all employing ``T5 Large'' as the PLM. The reported accuracies have been obtained (in December 2022) using the same random seed for consistency. The highest performing result is highlighted in bold.}
\label{tab:perform-comp}
\vspace{5pt}
\end{table}

\begin{figure}
    \centering
    \includegraphics[width=.44\textwidth,height=6cm]{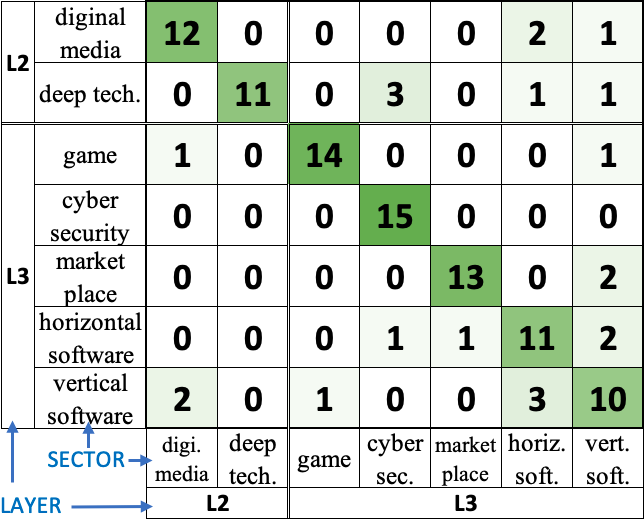}
    \caption{The confusion matrix for seven sectors picked from the 84 (as of December 2022) predicted sectors.}
    \label{fig:cm}
\end{figure}

Then we dig down to understand the error contribution from each sectors and find that sectors on low levels (e.g.,~L3 and L4 in Figure~\ref{fig:sf}) have an accuracy of over 90\% except two L3 sectors named {\it horizontal software} and {\it vertical software}, as reflected in Figure~\ref{fig:cm}.
A horizontal software company caters to a wide and broad ranging market of consumers, and a vertical one provides a solution for a particular line of business or industry.
Because of the way they are defined, many businesses in horizontal/vertical software sector might fit in other sectors as well.
%Although they are defined differently, many businesses are part of both horizontal and vertical software sectors simultaneously.
For example, a company providing bot-based customer service could be part of the horizontal market of any customer support scenario, while also targeting vertically to game publishers.
The complete list of predicted sectors is considered to be sensitive proprietary information and therefore we only show the confusion matrix for seven sectors (two from L2 and five from L3) in Figure~\ref{fig:cm}. 
Since L3 sectors are more fine-grained requiring less (than L2) annotations, a generally better inference performance is observed for L3 than L2, which encourages us to run a bottom-up annotation attribution (cf.~Section~\ref{sec:aa}) to prioritize lower-level sectors.

\begin{figure}
    \centering
    \includegraphics[width=.47\textwidth]{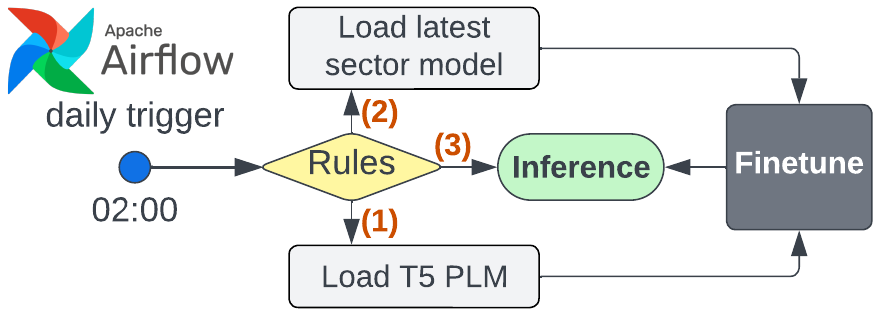}
    \caption{The full system diagram with three scenarios (1$\sim$3) controlled by a rule-based inspection operator.}
    \label{fig:sd}
\end{figure}

\section{The Full System}
The evolving annotation ({\bf\small Chall.1}) and dynamic sector framework ({\bf\small Chall.2}) both demand constant model iteration;
and any model update would require a full-scale re-inference.
As a result, triggering model iteration upon any change in annotation or sector framework will be computationally expensive and hard to scale.
Instead, We trigger a rule-based inspection only once every day (e.g.,~at about {\small \verb|02:00|} as exemplified in Figure~\ref{fig:sd}) through Airflow\footnote{Apache Airflow: \url{https://airflow.apache.org}}.
The inspection rules lead to three scenarios:
\begin{enumerate}[label={(\arabic*)}]
  \item {\bf Finetune on T5 PLM} when the sector framework is changed or the annotation for any existing sector has evolved significantly\footnote{For sector $\mathsf{s}_m$ that originally has $|\mathsf{C}_m|$ annotated companies, the number of newly added/removed companies is $\Delta_m$, then $\Delta_m/|\mathsf{C}_m|\geq0.75$ is regarded as {\bf significant}, and $0.75>\Delta_m/|\mathsf{C}_m|\geq0.1$ is a {\bf marginal} change.}; it takes about 7 hours on 2 $\times$ Nvidia P100 GPU.
  \item {\bf Finetune on the latest sector model} when the sector annotation only changed marginally$^{\text{9}}$; but the first scenario will be enforced after 90 days since its last execution.
  \item {\bf Skip finetune} otherwise and run incremental inference introduced in Section~\ref{sec:inf}.
\end{enumerate}
The second scenario takes less than 1/7 of the effort of the first scenario.
We continue to present the key ingredients of finetune and inference.

\subsection{Finetune}
\begin{figure}
    \centering
    \includegraphics[width=.485\textwidth]{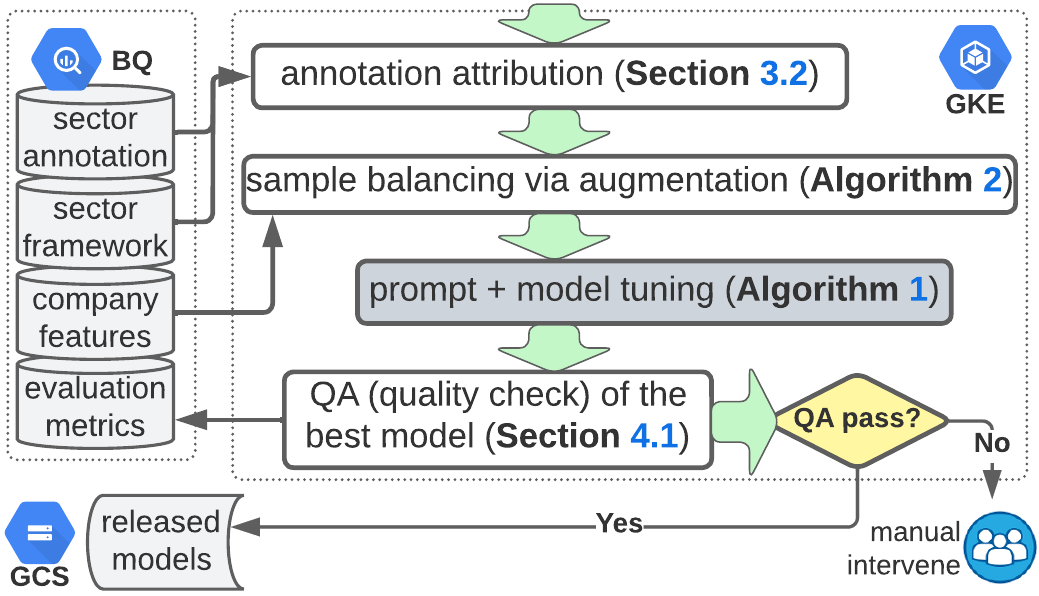}
    \caption{The finetune workflow. {\small BQ: BigQuery, GCS: Google Cloud Storage, GKE: Google Kubernetes Engine}.}
    \label{fig:ft}
\end{figure}

Figure~\ref{fig:ft} shows the finetune pipeline which is encapsulated in a docker\footnote{\url{https://www.docker.com}} image run by Google Kubernetes Engine (GKE)\footnote{\url{https://cloud.google.com/kubernetes-engine}}.
From our data warehouse managed by BigQuery\footnote{\url{https://cloud.google.com/bigquery}} \cite{melnik2010dremel}, the annotation attribution (Section~\ref{sec:aa}) collects all eligible sectors $\mathsf{s}_1,\ldots,\mathsf{s}_M$ together with their corresponding company sets $\mathsf{C}_1,\ldots,\mathsf{C}_M$, which are balanced via augmentation (Algorithm~\ref{algo:sba}).
The balanced dataset is then split (with a ratio of 9:1) into training and validation sets that are used for prompt + model tuning following Algorithm~\ref{algo:pmt}.

The validation metrics (sector-wise precision and recall calculated from a confusion matrix like Figure~\ref{fig:cm}) of the finetuned model will go through a QA (quality check) step to determine if this model is good enough to be automatically released in GCS (Google Cloud Storage)\footnote{\url{https://cloud.google.com/storage}}.
QA constitutes a series of assertions such as ``{\small\it The {\bf precision} of {\bf vertical software} should be greater than {\bf 75\%}}'' and so on.
If any of these assertions fails, it will send an alarm to our data scientists via Slack\footnote{\url{https://slack.com}} to request a manual interference to take appropriate actions. 

\begin{figure}
    \centering
    \includegraphics[width=.485\textwidth]{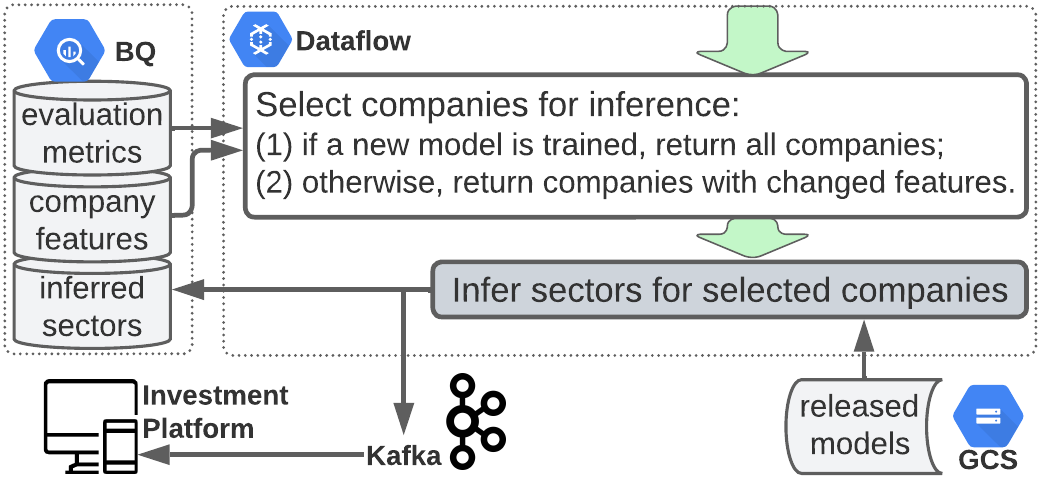}
    \caption{The (differentiated) inference workflow that starts with a rule-based company selection step. 
    %{\small BQ: BigQuery, GCS: Google Cloud Storage, Dataflow is suitable for job scaling.}
    }
    \label{fig:inf}
\end{figure}

\subsection{Inference}
\label{sec:inf}
The inference workflow starts with a selection step (cf.~Figure~\ref{fig:inf}) to determine a subset of companies that need re-inference. 
The selection step greatly reduce the daily inference load (by 95\% at least after the system stabilizes), hence it tackles {\small\bf Chall.4}.
Next, the latest trained model is loaded to infer the sectors for the selected companies,
where two facts could relieve {\small\bf Chall.4} further: (1) being able to use a medium-sized PLM, and (2) job parallelization by Dataflow\footnote{\url{https://cloud.google.com/dataflow}}.
The inferred sectors are stored in BigQuery and simultaneously published to Kafka\footnote{\url{https://kafka.apache.org}} so that our investment platform can further leverage those predictions in many PE analytical use cases.

\section{Related Work}
\label{sec:rw}
As discussed in Section~\ref{sec:model}, the most relevant approach is $M$-way classification using either word or sentence level features as input.
But the features are usually pre-learned with a fundamentally different setup and target (e.g.,~MLM: Masked Language Model, cf.~\citealp{devlin-etal-2019-bert}), which makes it potentially difficult to continue finetuning towards a classification target \cite{gururangan-etal-2020-dont}.
There is a recent trend of unifying all downstream tasks as a text generation problem \cite{lester-etal-2021-power}, i.e., a generative NLP paradigm.
In fact, using prompt has become the symbol of this paradigm.
For example, we can potentially ask a GPT-3 \citealp{brown2020language} or InstructGPT \cite{ouyang2022training}: ``{\small\it Klarna is a company that provide an cashless online payment platform. What is Klarna's industry sector?}''
The likely answer would be something that generally make sense, yet will not be mapped directly towards the predefined sector framework.
As a result, prompting \cite{liu2021pre} and prompt tuning \cite{su2022transferability} emerge to fill this gap.
However, \citet{lester-etal-2021-power} discover that model tuning still prevails when the size of PLM is relatively small, which inspires us to jointly tune small PLM and prompt, as explained in Section~\ref{sec:prompt_model_tuning}.

\section{Conclusion}
In order to support thematic PE fund operations, we design and deploy a scalable and adaptive system to infer customized industry sectors for millions of companies.
We empirically show that a generative NLP model is superior to its discriminative counterpart, leading to a solution of model + prompt tuning that guarantees superior performance even using scarce annotation and medium-sized PLM.
The prompt template is designed to cope with noisy input textual features.
To address the ever-changing sector framework and annotation,
the system automatically triggers and determines the most appropriate scenario by quantifying the change.
Moreover, the system also incorporates best-effort annotation attribution, sample balancing, and incremental inference. 
Hundreds of PE professionals has benefited from this system for over a year.
%Future work is calibrating the sector inference with company knowledge graphs.
Last but not least, our solution can be directly generalized to many similar scenarios such as e-commerce product tagging.

\section*{Acknowledgements}
We are grateful to the support from the entire \href{https://eqtgroup.com}{EQT} organization.
We also thank the constructive feedbacks from the reviewers of FinNLP@IJCAI 2023.
This work is also reviewed by EQT's compliance, communication and legal department prior to publication.

% Entries for the entire Anthology, followed by custom entries
\bibliography{custom}
\bibliographystyle{acl_natbib}

\end{document}